\documentclass[10pt,journal,compsoc]{IEEEtran}

\usepackage{amsmath}
\usepackage{nccmath}
\usepackage{graphicx}
\usepackage{rotating}
\usepackage{tikz}
\usepackage[normalem]{ulem}
\usepackage{color}
\usepackage{multirow}
\usepackage{hhline}
\usepackage{ragged2e}

\DeclareMathOperator*{\argmin}{argmin}

\ifCLASSOPTIONcompsoc
  \usepackage[nocompress]{cite}
\else
  \usepackage{cite}
\fi

\begin{document}
\title{Building Robust Deep Neural Networks for Road Sign Detection}

\author{Arkar Min Aung,Yousef Fadila, 
        Radian Gondokaryono,
        Luis Gonzalez}
        
% \IEEEcompsocitemizethanks{\IEEEcompsocthanksitem M. Shell was with the Department
% of Electrical and Computer Engineering, Georgia Institute of Technology, Atlanta,
% GA, 30332.\protect\\
% % note need leading \protect in front of \\ to get a newline within \thanks as
% % \\ is fragile and will error, could use \hfil\break instead.
% E-mail: see http://www.michaelshell.org/contact.html
% \IEEEcompsocthanksitem J. Doe and J. Doe are with Anonymous University.}% <-this % stops a space
% \thanks{Manuscript received April 19, 2005; revised August 26, 2015.}}

\markboth{Building Robust Deep Neural Networks for Road Sign Detection}%
{Shell \MakeLowercase{\textit{et al.}}: Bare Advanced Demo of IEEEtran.cls for IEEE Computer Society Journals}

\IEEEtitleabstractindextext{%

\justify
\begin{abstract}
Deep Neural Networks are built to generalize outside of training set in mind by using techniques such as regularization, early stopping and dropout. But considerations to make them more resilient to adversarial examples are rarely taken. As deep neural networks become more prevalent in mission critical and real time systems, miscreants start to attack them by intentionally making deep neural networks to misclassify an object of one type to be seen as another type. This can be catastrophic in some scenarios where the classification of a deep neural network can lead to a fatal decision by a machine. In this work, we used GTSRB dataset to craft adversarial samples by Fast Gradient Sign Method \cite{goodfellow2014explaining} and Jacobian Saliency Method \cite{papernot2016limitations}, used those crafted adversarial samples to attack another Deep Convolutional Neural Network and built the attacked network to be more resilient against adversarial attacks by making it more robust by Defensive Distillation \cite{papernot2016distillation} and Adversarial Training \cite{goodfellow2014explaining}.
\end{abstract}

}

\maketitle

\IEEEdisplaynontitleabstractindextext
\IEEEpeerreviewmaketitle

\section{Introduction}
\label{sec:introduction}
With the availability of more computational resources and abundance of data, there has been a huge resurgence of using deep neural networks to do object recognition and classification but several machine learning models, including state-of-the-art deep neural networks, consistently misclassify adversarial examples, which are inputs formed by applying small, but intentionally engineered, worst-case perturbations to input images. These perturbations are indiscernible for humans, but they can make deep neural networks to make wrong classifications with very high confidence. The problem becomes more concerning with the advent of self-driving cars which does automatic detection and classification of road signs to do path planning, adjusting speed or driving behaviors. If the Convolutional Neural Network which detects road signs in a self-driving car is fed with adversarial inputs, even though it is obvious for a human to classify it correctly, the network may make an egregious misclassification of that road sign. This can result in self-driving cars making erroneous decisions.

In this work, ways to create adversarial examples from road sign images are explored in order to use them to fool the state-of-the-art neural networks and an effort to build more robust neural networks to be resilient against these attacks is made. In Section \ref{sec:related}, some of the previous work that has been done related to adversarial examples is addressed. Explanations of the methods that were used to craft adversarial examples and the ways used to build more robust neural networks to be resilient against adversarial samples are presented in Section \ref{sec:methodology}. The dataset used and the data augmentation processes are also described in \ref{sec:methodology}. Experimental results are shown in Section \ref{sec:result} and finally, further discussions on the weakness of this work as well as the possible future extensions of this work are discussed in Section \ref{sec:discussion}. Finally, the scope of the work is concluded in Section \ref{sec:conclusion}.

\section{Related Work}
\label{sec:related}
The work by Nguyen et al. \cite{nguyen2015deep} was the inception of fooling state-of-the-art neural networks. In this work, it is shown that deep neural networks are easily fooled to classify images which are not recognized by humans as 
belonging to particular classes with high confidence. \cite{szegedy2013intriguing} first showed how state-of-the-art machine learning models, including neural networks, are vulnerable to adversarial examples. Based on the work of \cite{szegedy2013intriguing}, \cite{goodfellow2014explaining} presented how  to generate adversarial examples with Fast Gradient Sign Method, how adversarial training can result in further regularization than dropout and how adversarial examples generalize across different deep neural network models. \cite{papernot2016limitations} further explores the reason behind adversarial attacks and presented how imperfections in the training phase of deep neural networks can make them vulnerable to adversarial samples. \cite{papernot2016limitations} formalizes the space of adversaries against deep neural networks and introduces a novel class of algorithms to craft adversarial samples. \cite{papernot2016practical} addressed the issue of black-box adversarial attacks which aligns with the problem space that our work is aiming to address.
Not only are adversarial samples crafted by perturbing pixels in the image, \cite{kurakin2016adversarial} and \cite{evtimov2017robust} explored how adversarial examples can be transferred to the physical world, such as adversarial printed road signs with graffiti like art on top of it. These road signs seem like they are being vandalized but for neural networks those graffiti overlays lead the classification astray from the correct prediction. Nonetheless, recent work by \cite{lu2017no} and \cite{lu2017standard} stated that standard detectors such as FasterRCNN \cite{ren2015faster} and YOLO \cite{redmon2016yolo9000} are not fooled by physical adversarial stop signs.

Based on the method knowledge distillation described in \cite{hinton2015distilling}, Papernot et al proposed a method of making neural networks more resilient to adversarial samples in \cite{papernot2016distillation}. But \cite{carlini2016defensive} stated that defense against adversarial samples by 
defensive distillation does not work. Since, the field of adversarial sample crafting and defense against adversarial attacks is a fairly new research topic in machine learning, there has been various literature debating on the effectiveness of different methods. Our work serves as the connecting hub to tie up different ends of literature both on the attacking deep neural networks with adversarial samples, defending against adversarial attacks and verifying whether the defensive methods work on dataset different from datasets used in literature. 

\section{Methodology}
\label{sec:methodology}

\begin{figure}[tpb]
      \centering
	  {\includegraphics[width=\linewidth,keepaspectratio]
{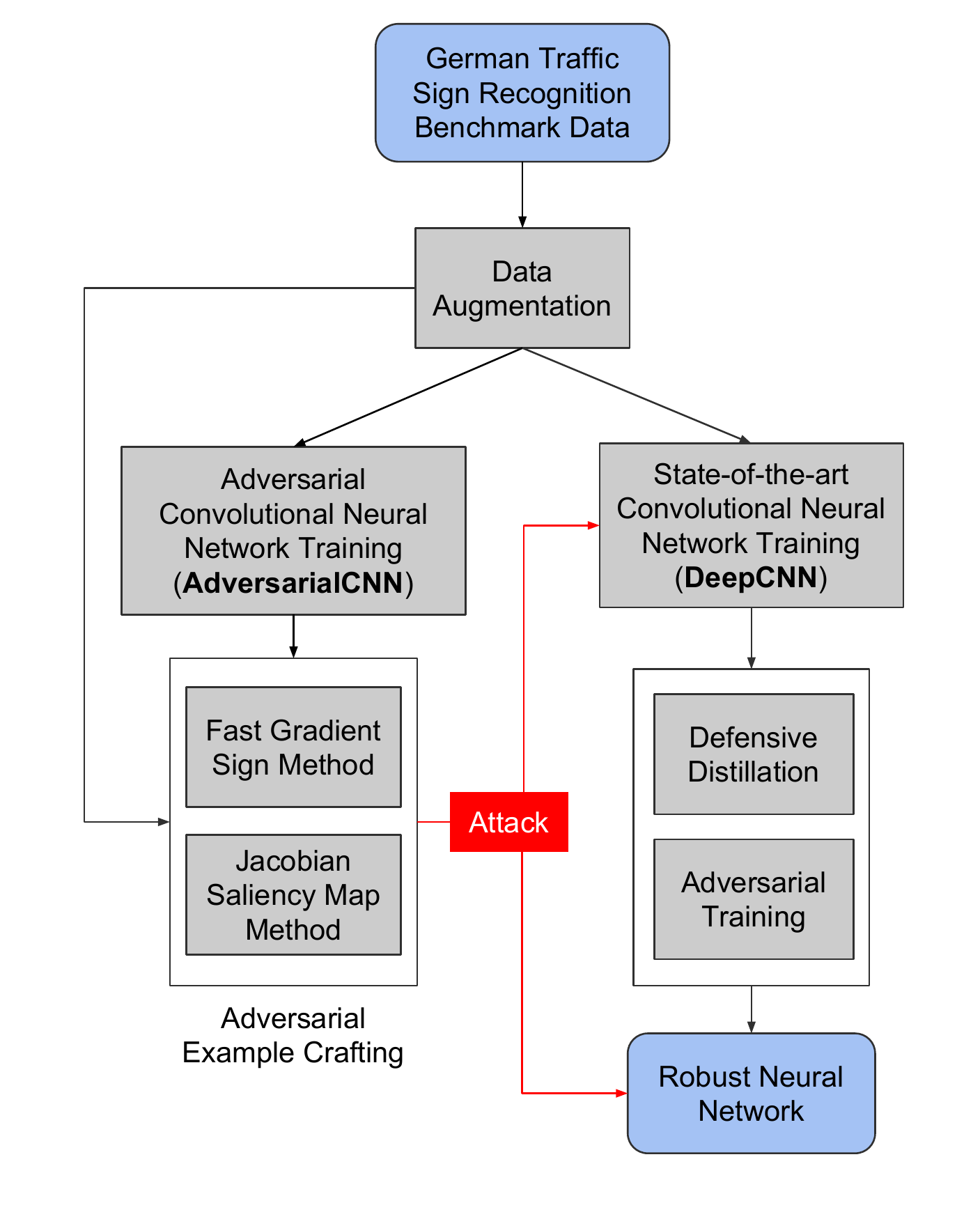}}
      \caption{Pipeline of methodology implemented in this work of crafting adversarial samples, attacking state-of-the-art convolutional neural network and building more robust neural network for road sign detection.}
      \label{fig:pipeline} 
\end{figure}

\subsection{Pipeline}
\label{sec:pipeline}
There are three main objectives in this work: building a state-of-the-art traffic sign classification deep convolutional neural network, crafting adversarial samples from GTSRB dataset, fooling the state-of-the-art network and making the network to be more resilient against the attacks. The methodology flowchart from Figure \ref{fig:pipeline} depicts the scope of creating a robust neural network built upon the objectives. 

The first objective, which is the first step in our pipeline, is to build a classifier which has near state-of-the-art performance on test set of GTSRB dataset. In order to satisfy this objective, we trained and tested a Convolutional Neural Network (which we will call \textbf{DeepCNN} from now on). The detailed architecture of DeepCNN is depicted in Table \ref{table2}. DeepCNN is trained for $30$ epochs with SGD optimizer with learning rate $0.01$, with momentum and learning rate decay. DeepCNN is the network which the adversary wants to attack but the architecture, weights and optimizers are not known by the adversary. 

The second objective, which is the second step in our pipeline, is to craft adversarial samples based on GTSRB dataset. These adversarial samples are used to attack DeepCNN, without knowing its architecture and weights. In order to craft adversarial examples, another CNN is built. We call this CNN, \textbf{AdversarialCNN}. Detailed architecture of AdversarialCNN is depicted in Table \ref{table1}. AdversarialCNN is used to craft adversarial samples using Fast Gradient Sign Method (Section \ref{sec:fgsm}) and Jacobian-based Saliency Map Method (Section \ref{sec:jacobian}). The adversarial samples generated are first analyzed whether the perturbations are visually perceptible and then the best hyperparameters which produce the least visually perceptible adversarial samples are picked to generate more samples  to attack the DeepCNN.

The final step in our pipeline of this work is to use two defense against adversarial attacks methods called Adversarial Training (Section \ref{sec:adversarialtraining}) and Defensive Distillation (Section \ref{sec:defensivedistillation}). These two methods are used to make the DeepCNN to be more robust against adversarial attacks. 

Both DeepCNN and AdversarialCNN are trained from scratch without their weights being initialization from pre-trained networks. The reason for training both from scratch is to emulate the hybrid Black Box attack as close as possible. Since both models were trained from scratch, each model does not have any knowledge of the other model. Therefore, AdversarialCNN is generating samples by only knowing the input dataset and the output classes of the GTSRB dataset. The reason that the AdversarialCNN still has the knowledge of the dataset restricts our approach from having a full Black Box attack.

The motivation behind building two CNNs with different architectures is as follows: assume that an adversary wants to attack a neural network which is doing a traffic sign classification in a self-driving car. The adversary does not know the architecture nor the weights of the neural network but knows the input and output pairs. For example, the attacker knows that given a stop sign image, the network predicts the class label of a stop sign. Therefore, the attacker will try to build a replica network with different architecture and try to generate adversarial samples which can potentially attack a deeper, more complex network in a black box fashion. If we compare the architecture of DeepCNN (Table \ref{table2}) and AdversarialCNN (Table \ref{table1}), it can be seen that DeepCNN is deeper and has more reprsentational power. The reason for making AdversarialCNN less complex and less powerful is to make the task of generating adversarial examples as hard as possible to attack a deeper and more complex model. This emulates a more realistic scenario where a deployed neural network is likely to be more complex when compared to a replica network which a miscreant would have built for attacking it.

\subsection{Dataset}
% TALK ABOUT WHY CLASS IMBALANCE IS NOT DESIRED
For this work, the German Traffic Sign Recognition Benchmark (GTSRB) \cite{Stallkamp-IJCNN-2011} dataset was used which is popular benchmark for deep learning problems. This dataset is composed of more than $20,000$ images, each image belonging to one of $43$ classes. The dataset is split into training set and testing set. Since we are building a Convolutional Neural Network with fixed input size, it was decided to work with RGB images of size 32x32. Furthermore, all of the images with a smaller size were removed.
Since there is a high class imbalance in the training samples which can be seen in histogram presented in Figure \ref{fig:data_augmentation}, a decision of generating more data for the classes with a low quantity of samples was taken.

\begin{figure}[tpb]
      \centering
	  {\includegraphics[width=\linewidth]{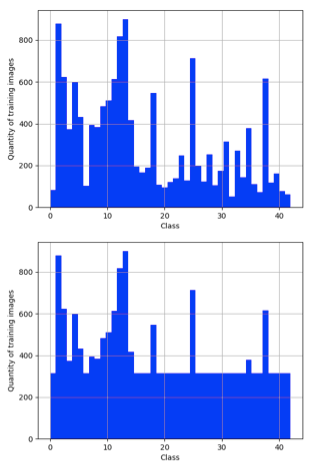}}
      \caption{Histogram of samples over $43$ classes. Class imbalance is visible in original dataset (Upper histogram). Histogram of samples over $43$ class after data augmentation (Lower histogram). The mean samples per class is chosen as the cutoff for augmenting samples in classes which are lower than the mean samples per class.}
      \label{fig:data_augmentation_hist} 
\end{figure}

\subsubsection{Data Augmentation}
It is well known that the more data machine learning algorithms have access to, the more effective they are. A common way of doing data augmentation is by performing of affine transformations \cite{wangeffectiveness} on each image. An affine transformation is any transformation that can be expressed as a matrix multiplication and a vector addition. They can be used to perform rotations, translations and scale operations on images so they are basically a relation between to images using the following 2x3 matrix:

\begin{equation}
M = 
\begin{bmatrix}
A & B
\end{bmatrix}
=
\begin{bmatrix}
a_{00} & a_{01} & b_{00} \\
a_{10} & a_{11} & b_{10}
\end{bmatrix}
\end{equation}

In order to transform a vector $[x,y]^{T}$ using $M$, the following must be done:

\begin{equation}
x_{trans} = A
\begin{bmatrix}
x \\
y
\end{bmatrix}
+ B = M
\begin{bmatrix}
x & y & 1
\end{bmatrix}^{T}
\end{equation}

\begin{figure}[tpb]
      \centering
	  {\includegraphics[width=\linewidth]{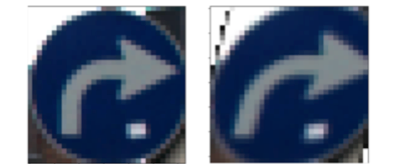}}
      \caption{Generation of a new training image by the application of an affine transformation}
      \label{fig:data_augmentation} 
\end{figure}

It is possible to generate new training examples by applying these transformations using $M$. Since it is infeasible to generate training data in a manual way, an automatic data augmentation process was done by applying small random perturbations to the base points and using these new ones in order to obtain different transformation matrices. An image generated by this process is shown in Figure \ref{fig:data_augmentation}.

% write about why data augmentation works
% why we need to 

\subsection{Network Architecture}
\label{sec:architecture}
The detailed architectures for AdversarialCNN and DeepCNN are described in Table \ref{table1} and Table \ref{table2}. Both networks take in 3-channel RGB image of width 32 pixels and height 32 pixels. AdversarialCNN is fairly shallow Convolutional Neural Network compared to DeepCNN. The main motivation behind intentionally making AdversarialCNN less complex than DeepCNN is depicted in Section \ref{sec:pipeline}. DeepCNN has a penultimate layer called Temperature layer with tunable parameter $T$. This layer is the implementation of the concept depicted in Section \ref{sec:defensivedistillation}.

\begin{table}[ph]
\centering
\caption{Network architecture for AdversarialCNN. This network is used to craft adversarial samples.}
\label{table1}
\begin{tabular}{|l|l|}
\hline
\multicolumn{1}{|c|}{\multirow{2}{*}{\textbf{Layer Name}}} & \multicolumn{1}{c|}{\textbf{Adversarial CNN}}                                                                                    \\ \cline{2-2} 
\multicolumn{1}{|c|}{}                                     & \multicolumn{1}{c|}{\textbf{Specification}}                          \\ \hline
Input & Shape: 32x32x3                                                                      \\ \hline
Conv 1                                                     & \begin{tabular}[c]{@{}l@{}}Kernel Size: 8x8\\ Number of filters: 64\\ Stride (x, y): (2, 2)\\ Padding: zero-padding\end{tabular} \\ \hline
Activation                                               & ReLU                                                                                                                                                                     \\ \hline
Conv 2                                                     & \begin{tabular}[c]{@{}l@{}}Kernel Size: 6x6\\ Number of filters: 128\\ Stride (x, y): (2, 2)\\ Padding: No Padding\end{tabular}  \\ \hline
Activation                                               & ReLU                                                                                                                             \\ \hline
Conv 3                                                     & \begin{tabular}[c]{@{}l@{}}Kernel Size: 5x5\\ Number of filters: 128\\ Stride (x, y): (1, 1)\\ Padding: No Padding\end{tabular}  \\ \hline
Activation                                               & ReLU                                                                                                                             \\ \hline
\multicolumn{2}{|c|}{Flatten}                                                                                                                                                                 \\ \hline
Dense 1                                                    & Nodes: 1024                                                                                                                      \\ \hline
Output                                                     & Nodes: 43                                                                                                                        \\ \hline
\end{tabular}
\end{table}

\begin{table}[tbp]
\centering
\caption{Network architecture for DeepCNN. This network is the network which is first attacked by adversarial samples and then made robust by defensive distillation and adversarial training. Note that the penultimate layer, Temperature Layer, has the key tunable parameter $T$ to get entropic softmax outputs.}
\label{table2}
\begin{tabular}{|l|l|}
\hline
\multicolumn{1}{|c|}{\multirow{2}{*}{\textbf{Layer Name}}} & \multicolumn{1}{c|}{\textbf{Deep CNN}}                                                                                           \\ \cline{2-2} 
\multicolumn{1}{|c|}{}                                     & \multicolumn{1}{c|}{\textbf{Specification}}                         \\ \hline
Input & Shape: 32x32x3                                                                       \\ \hline
Conv 1                                                     & \begin{tabular}[c]{@{}l@{}}Kernel Size: 3x3\\ Number of filters: 32\\ Stride (x, y): (1, 1)\\ Padding: Zero Padding\end{tabular} \\ \hline
Activation                                                 & ReLU                                                                                                                             \\ \hline
Conv 2                                                     & \begin{tabular}[c]{@{}l@{}}Kernel Size: 3x3\\ Number of filters: 32\\ Stride (x, y): (1, 1)\\ Padding: Zero Padding\end{tabular} \\ \hline
Activation                                                 & ReLU                                                                                                                             \\ \hline
Max Pool 1                                                 & \begin{tabular}[c]{@{}l@{}}Window Size: 2x2\\ Stride (x, y): (2, 2)\end{tabular}                                                 \\ \hline
Dropout 1                                                  & Probability: 0.25                                                                                                                \\ \hline
Conv 3                                                     & \begin{tabular}[c]{@{}l@{}}Kernel Size: 3x3\\ Number of filters: 64\\ Stride (x, y): (1, 1)\\ Padding: Zero Padding\end{tabular} \\ \hline
Activation                                                 & ReLU                                                                                                                             \\ \hline
Conv 4                                                     & \begin{tabular}[c]{@{}l@{}}Kernel Size: 3x3\\ Number of filters: 64\\ Stride (x, y): (1, 1)\\ Padding: Zero Padding\end{tabular} \\ \hline
\multicolumn{1}{|c|}{Activation}                           & \multicolumn{1}{c|}{ReLU}                                                                                                        \\ \hline
Dropout 2                                                  & Probability: 0.25                                                                                                                \\ \hline
\multicolumn{2}{|c|}{Flatten}                                                                                                                                                                 \\ \hline
Dense 1                                                    & Nodes: 256                                                                                                                       \\ \hline
Dropout 3                                                  & Probability: 0.5                                                                                                                 \\ \hline
Dense 2                                                    & Nodes: 43                                                                                                                        \\ \hline
Activation                                                 & ReLU                                                                                                                             \\ \hline
Temperature Layer                                          & Temp: T                                                                                                                          \\ \hline
Output                                                     & 43                                                                                                                               \\ \hline
\end{tabular}
\end{table}

%layers = [conv_2d(nb_filters, (8, , (2, 2), "same",
%                       input_shape=input_shape),
%               Activation('relu'),
%               conv_2d((nb_filters * 2), (6, 6), (2, 2), "valid"),
%               Activation('relu'),
%               conv_2d((nb_filters * 2), (5, 5), (1, 1), "valid"),
%               Activation('relu'),
%               Flatten(),
%               Dense(nb_classes)]

\subsection{Adversarial Example Crafting}

 An adversarial sample is an input crafted to impact deep learning algorithms output integrity. That could be an input that is deliberately built to cause the Artificial Neural Network to misclassify the input to a different class than what it is supposed to be or at least to reduce the output confidence. Such attacks could be targeted or untargeted. Targeted attack aims to cause a particular output class for an input while untargeted attack aims to misclassify the output to any other class regardless what is it. Crafting adversarial samples do not require any modification of the training process as they are created after the model has been fully trained.

The linear structure of most neural networks suggests weakness to linear perturbations in the images \cite{goodfellow2014explaining}. Equation \ref{pert} explains this phenomenon on how a small perturbation $\eta$ plus the original image $x$ affects the predictions of a classifier $w^T\tilde{x}$. A larger sized vector results a substantial change in the classifier. 

\begin{equation}
w^T\tilde{x} = w^Tx + w^T\eta
\label{pert}
\end{equation}

Adversarial attacks capabilities could be grouped into five categories defined by the information held by the adversary \cite{papernot2016practical}. 

\begin{enumerate}
	\item Training data and network architecture are known to the adversary
	\item Network architecture is known to the adversary but not the training data
	\item Training data is known to the adversary but not the network architecture
	\item Oracle mode: neither training data nor network architecture is known to the adversary but the adversary has access  to the model as an “oracle”. The adversary can get output from supplied inputs.
	\item Samples mode: the adversary has only a collection of pairs of input and output related to the classifier. 
\end{enumerate}

\subsubsection{Fast Gradient Sign Method}
\label{sec:fgsm}
Fast gradient sign method (FGSM) is a mathematical method of generating adversarial examples by derivation of parameters of the neural network \cite{goodfellow2014explaining}. Equation \ref{eq_fgsm} formulates this perturbation by taking the derivative of the cost function $J(\theta,x,y)$ to find the direction of the perturbation with a nominal size of $\epsilon$. This cost function is the confidence of label $y$ of image $x$. The resulting perturbations $\nu$ are added as in equation \ref{pert}. Let equation \ref{eq_d} be defined as the gradient of an image at a certain pixel $i_x, i_y$. 

\begin{equation}
\eta=\epsilon*sign(\nabla*J(\theta,x,y))
\label{eq_fgsm}
\end{equation}

\begin{equation}
d(\theta,x,y,i_x,i_y)=sign(\nabla*J(\theta,x,y,i_x,i_y))
\label{eq_d}
\end{equation}

\begin{figure}[tpb]
      \centering
	  {\includegraphics[width=\linewidth]{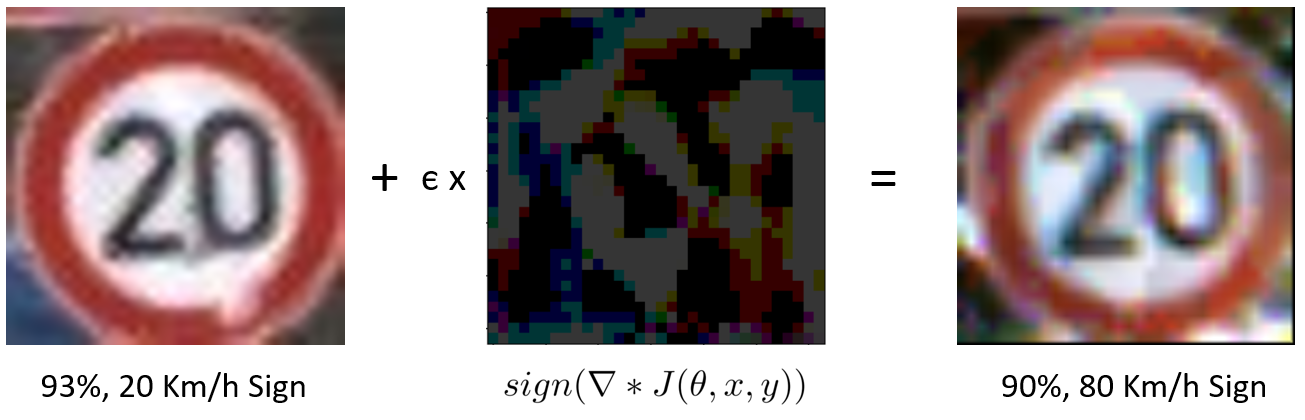}}
      \caption{Fast gradient sign method example of a 20 km/h road sign with perturbations misclassified as an 80 km/h road sign. The Perturbation image is normalized amplified data.}
      \label{fgsm_ex_fig} 
\end{figure}

Figure \ref{fgsm_ex_fig} shows an image example on how equations \ref{pert} and \ref{eq_fgsm} generates an adversarial example that misclassifies the 20 km/h road sign to an 80 km/h road sign. By examination of the adversarial example itself, a user would have difficulties discerning whether this is an adversarial image. Some images of different $\epsilon$'s (in section \ref{sec:result}) are perceptible adversarial examples. 

User perceptibly is defined as the ability to perceive an adversarial attack on an image. For FGSM, user perceptibly depends on the size of the feature vector and the choice of the parameter $\epsilon$. Here the feature vector is the image dimension n x n. Imperceptible adversarial examples are easier to craft from larger dimension images.

\subsubsection{Jacobian-based Saliency Map Method}
\label{sec:jacobian}

\begin{figure}[tpb]
      \centering
	  {\includegraphics[width=\linewidth]{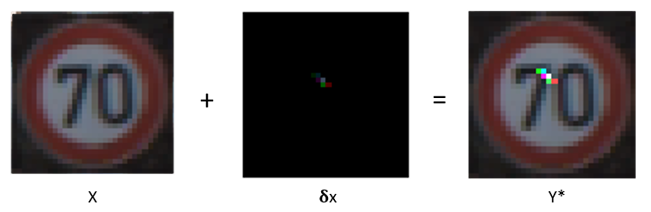}}
      \caption{Jacobian saliency map method example of a 70 km/h speed limit sign with minimally added perturbations misclassified as a 30km/h speed limit sign. }
      \label{jsma_ex} 
\end{figure}

Jacobian-based saliency map method was first introduced by Papernot et al. \cite{paper2016jsma}. This method is based on identifying small set of pixels candidates for perturbation in order to perform a class targeted attack. Equation \ref{eq:jacobian} describes how Jacobian-based Saliency works. Let $F:X \Rightarrow Y$ be the function learned from a trained neural network, where $X$ is a legitimate sample, $Y^*$ is a targeted class and $\delta_x$ is the small perturbation added to $X$. The objective is to perform optimization where $\delta_x$ is minimized and still satisfying the Equation \ref{eq:jacobian}. Figure \ref{jsma_ex} illustrates this example of adding a perturbation $\delta_x$ to the original image $x$.

\begin{equation}
\label{eq:jacobian}
\argmin_{\delta_{\textbf{X}}} \| \delta_{\textbf{X}} \| \quad \textbf{s.t.} \quad \textbf{F} (\textbf{X} + \delta_{\textbf{X}}) = \textbf{Y}^{*}
\end{equation}

This entails calculating the forward derivative of a neural network in order to build a saliency map that distinguishes the potential features for perturbation. Such perturbation would leads to the desired adversarial output. This approach is of category 2 attack (the adversary needs to know the network architecture and weight parameters to craft adversarial samples). In our research, we were able also to produce a category 3 attack  (training data is known but not the architecture) with high rate of success by generating the adversarial samples on different (and simpler) architecture.

To summarize this approach is processed is processed through these  four steps: 
\begin{enumerate}
	\item Compute the Jacobian matrix of $F$ evaluated at input $X$.
	\item Use Jacobian to find which features of input should be perturbed .
	\item Create $X^*$ by perturbing the features found in Step 2 on $X$.
	\item Repeat while $X^*$ is not misclassified and perturbation is still small.
\end{enumerate}

\subsubsection{Using Jacobian-based Saliency Map Approach to Craft Category 3 Attacks}

As we have mentioned, Jacobian-based saliency map approach assumes knowing the network architecture and weight parameters in order to build the saliency map that identifies the set of input features candidate for perturbation (category 2 attack). In our research, we show that even without knowing the network architecture we were able to craft adversarial samples if we has access to the training data (Category 3 attack). Even though the attacker does not know the architecture and weights of the network he wants to attack, the attacker can build a replica network which does the same classification task. The attacker would then be able to craft adversarial samples using his knowledge in his own network architecture and weights.
%In the context of our research, the defender architecture differs from the adversarial assumed architecture and yet $78\%$ of the adversarial samples generated on the adversarial network, were misclassified on the defender network too. % $62\%$ of them were misclassified to the same targeted class as the adversarial network, while the other $38\%$ were misclassified to a different class rather the original targeted class. %

\subsection{Building Robust Neural Networks}
\subsubsection{Defensive Distillation}
\label{sec:defensivedistillation}

Defensive distillation smooths the model’s decision surface in adversarial directions exploited by the adversary \cite{papernot2016distillation}. The idea behind defensive distillation is based on knowledge distillation which was first coined by Hinton et al. \cite{hinton2015distilling}. Knowledge distillation is a training procedure where one model is trained to produce entropic outputs using the original one-hot encoded training labels and the second model uses those entropic soft target outputs as the training label. In the original work, the main motivation behind knowledge distillation is to distill the knowledge from bigger network to a smaller network which can be deployed on mobile devices but still having the equivalent representative power of a bigger network. 

In the case of defensive distillation, there is no need to have a smaller model as the main goal is not compression but more robust networks. Therefore, in the setting of defensive distillation, the network of same architecture is trained twice; first, with the original one-hot encoded labels to produce entropic predictions by tuning a temperature parameter (described in next paragraph) and second, using those entropic soft targets as training labels which are fed back into the same network and retrained from scratch with new labels. Distillation makes the final model's responses smoother, and therefore, it works even if two models are of the same size. Using the same network architecture to product the outputs which will be fed back to the same network for training may sound counterintuitive but the reason it works is that the first model is trained with “hard” labels ($100\%$ probability that an image is a 80km/h road sign rather than a stop sign) and then provides “soft” labels ($70\%$ probability that an image is a 80km/h, $20\%$ probability that an image is a 60km/h, etc) which are used to train the second model. This implicitly makes the network to be less confident of its predictions. Having a network with less confident predictions is useful when the network is making wrong predictions. It is more desirable to have network with less confident wrong predictions than a network with very confident wrong predictions. Even if the network is making wrong predictions, it is better to make wrong predictions which are semantically similar to the actual class or have the actual class showing up as second or third top prediction. Figure \ref{fig:softmaxpreds} (a) and (b) shows the resulting predictions of a confident network (i.e, network trained without defensive distillation). Figure \ref{fig:softmaxpreds} (c) and (d) shows the resulting predictions of a less confident network. The second and third predictions of the less confident network is more semantically related to the actual class label compared to the second and third predictions of a very confident classifier. Therefore, in the setting of adversarial sample crafting, small tugs of perturbations can lead the network to make confident incorrect classifications if the adversary can find the sharp edges in the manifold of the class boundaries.

\textbf{Temperature}: To perform distillation, a Convolutional Neural Network network whose output layer is a softmax is first trained on the original dataset. The description of the original network is described in Section \ref{sec:architecture}. Consider $Z(X)$, the \emph{logit} outputs produced by the last hidden layer of the CNN, right being trasformed to normalized probabilities with \emph{softmax} function $F(X)$. The outputs obtained from $F(X)$ describe the probability how likely that data $X$ is  class $N$. Within the \emph{softmax} layer, a given neuron corresponding to a class indexed by $i \in 0..N-1$ (where $N$ is the number of classes, $43$ in our case) computes the component $i$ of the following output vector $F(X)$:

\begin{equation}
F(x) = \Bigg[ \frac{e^{z_i(X)/T}}{\sum_{l=0}^{N-1}e^{z_i(X)/T}} \Bigg]_{i \in 0...N-1 }
\end{equation}

The parameter $T$ is a tunable parameter and can be implemented as an additional layer between \emph{softmax} and final fully connected layer of the size $N$ nodes. This additional layer only needs to perform element-wise division to the output row vector from the fully connected layer.

\subsubsection{Adversarial Training}
\label{sec:adversarialtraining}

Using adversarial examples generated with Fast Gradient Sign Method and Jacobian Saliency Map Method, we split the training adversarial set and testing adversarial set which is the same as GTSRB dataset train and test split. Training adversarial samples are fed back in the neural network as new training samples. Adversarial training is similar to brute force approach of solving more robust neural networks but generating more adversarial samples are expensive and not is not scalable. Therefore, adversarial training is used as a supplement to help defensive distillation which is more scalable since it only needs one parameter (temperature $T$) to be tuned.

\section{Results}
\label{sec:result} 
Experimental results based on stated methodology are obtained and discussed separately each for adversarial crafting, attack on DeepCNN, Defensive Distillation and Adversarial Training.

\subsection{Adversarial Examples: Fast Gradient Sign Method}
	The adversarial neural network described in section \ref{sec:architecture} was trained with the GTSRB data. The resulting accuracy of the network is $93.49\%$. By back-propagation of the network, we initiate the fast gradient sign method attack with help of the Cleverhans library. \cite{papernot2017cleverhans} 23,000 adversarial examples were generated for each changing parameter $\epsilon$ (eq. \ref{eq_fgsm}) 0.01 to 0.30 with a step of 0.01. 
   
\begin{figure}[tpb]
      \centering
	  {\includegraphics[width=\linewidth]{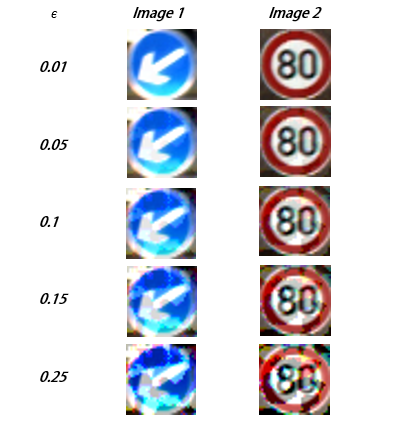}}
      \caption{Adversarial example images from fast gradient sign method with increasing parameters $\epsilon$. Pixel change pattern is similar for each image. Magnitudes of the perturbation increase with increasing $\epsilon$.} 
      \label{fgsm_eps}
\end{figure}

   Figure \ref {fgsm_eps} shows the resulting adversarial examples generated from 2 images with increasing $\epsilon$. The last column explains whether the image has been either classified or missclassified. We can see that increasing $\epsilon$ changes the magnitude of perturbation as well as increasing the chances for misclassification of the image. Let $d(\theta, x, y, i_x, i_y)$ be the direction of perturbation of each pixel $i_x, i_y$ as defined in equation \ref{eq_d}. For each changing $\epsilon$ in figure \ref{fgsm_eps}, we can observe that the direction of perturbation $d(\theta,x,y,i_x,i_y)$ is constant. In other words, the perturbed image retained a constant pattern through each iteration $\epsilon$.

    Perceptibility means whether a user can classify the image as an attack on the neural network. The result is dependent on the image itself and the subjectivity of the person viewing the image. Compare images 1 and 2 for $\epsilon= 0.1$. Qualitatively by the authors assessment , the image on the left is perceptible and the image on the right is imperceptible. 
     
\begin{figure}[tpb]
      \centering
	  {\includegraphics[width=\linewidth]{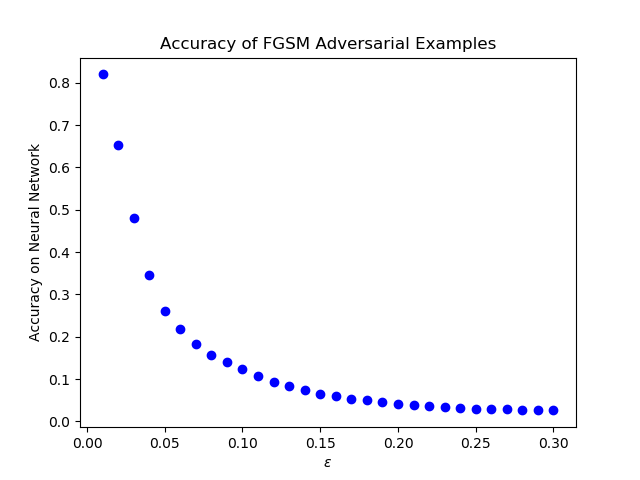}}
      \caption{Accuracy vs $\epsilon$ of adversarial examples from fast gradient sign method. For each $\epsilon$, accuracy is tested on the AdversarialCNN with 23,000 adversarial examples. Increasing $\epsilon$ exponentially decays the accuracy of the neural network} 
      \label{eps_acc}
\end{figure}

     Another result to consider is shown in Figure \ref{eps_acc} which plots the accuracy of the neural network with increasing epsilons. Here accuracy is defined based on the number of images correctly classified from the total of images tests. We can see an exponential decay of accuracy with increasing $\epsilon$'s. 

\subsection{Adversarial Examples: Jacobian Saliency Map Method}
\begin{figure}[tpb]
      \centering
	  {\includegraphics[width=\linewidth]{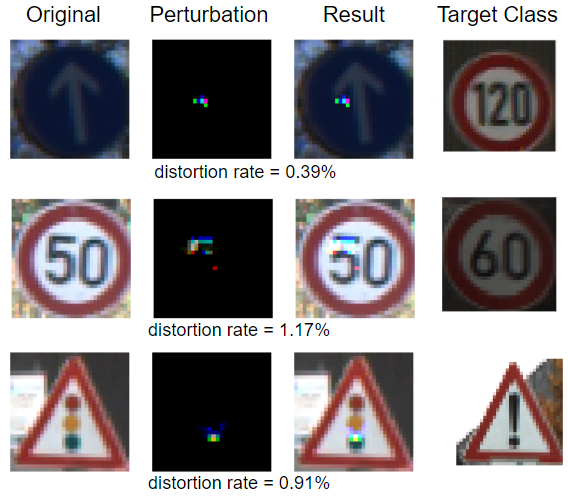}}
      \caption{Examples of successfully generated targeted adversarial samples using Jacobian  saliency map method}
      \label{jsma_pairs_fig} 
\end{figure}

	The simple neural network to generate adversarial examples uses parameters described in section \ref{sec:architecture} trained using the GTSRB data, three channels RGB. In order to generate adversarial samples, we first chose 43 samples, one of each class, and then for each one, we had tried to make a targeted attack to all other classes. In total we have generated $43 * 42 = 1806$ samples. The success rate, which is percentage of adversarial samples that were successfully classified by the DNN as the adversarial target class, was $82\%$. The distortion, which is percentage of pixels modified in the legitimate sample to obtain the adversarial sample was $1.74\%$ in average. Figure \ref {jsma_pairs_fig} shows three examples of successfully generated targeted adversarial samples

\subsection{Deep CNN and Attack on DeepCNN}

During the initial training step of DeepCNN, temperature parameter ($T$) is set to $1$. When temperature parameter is set to $1$ the Temperature layer (Penultimate layer in Table \ref{table2}) is analogous to identity mapping. Moreover, the network is not restricted to produce entropic softmax predictions and therefore, after $20$ epochs, the network becomes very confident with its predictions. 

\begin{figure}[tpb]
\label{fig:softmaxpreds}
      \centering
	  {\includegraphics[width=\linewidth]{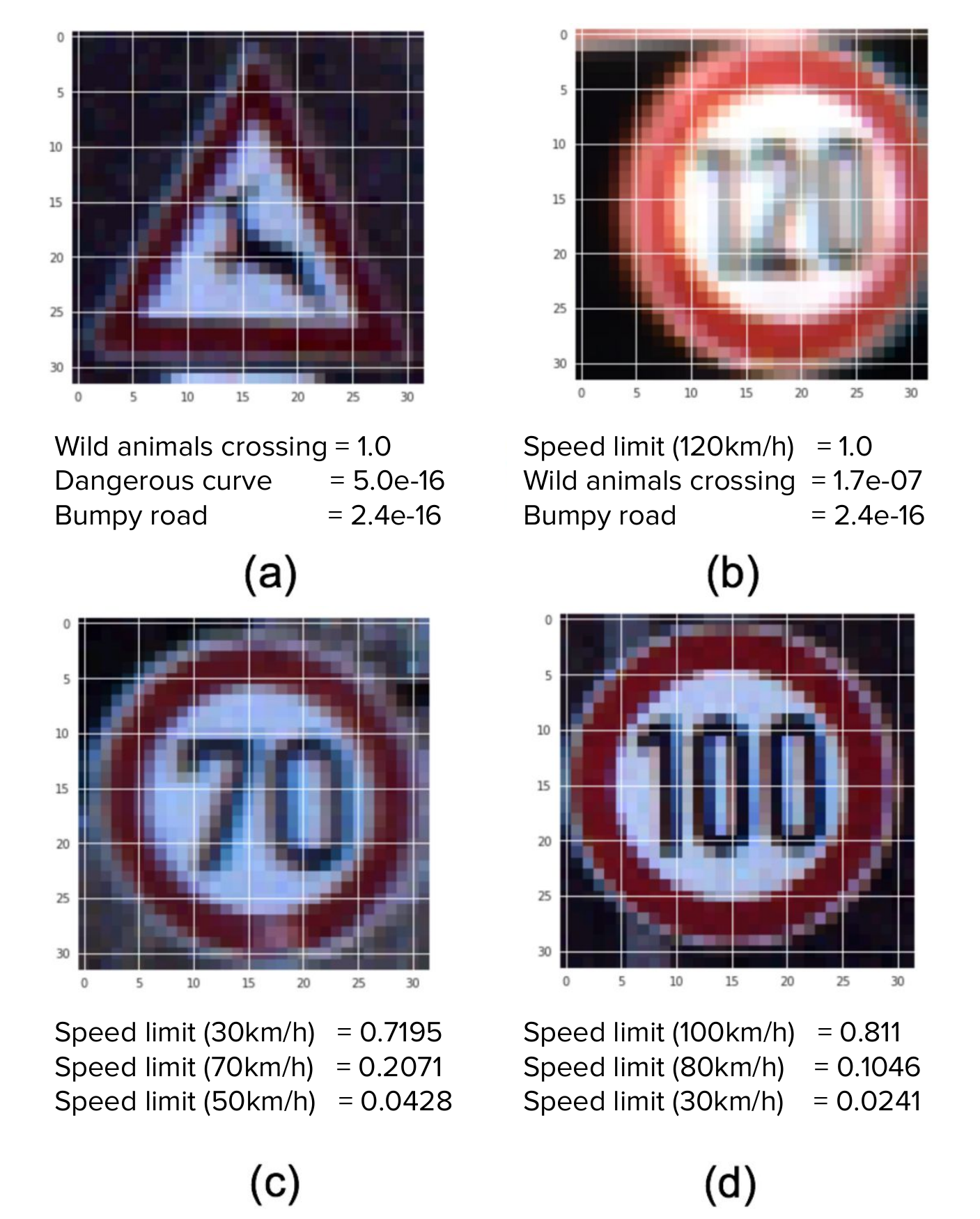}}
      \caption{Examples of confident predictions and less confident predictions. (a) and (b) are produced by DeepCNN with $T=1$. Networks with high confidence tend to make confident correct predictions as well as confident wrong predictions. (c) and (d) are produced by DeepCNN with $T=100$. Networks with low confidence tend to make less confident correct predictions as well as less confident wrong predictions. Even though the first prediction in (c) is wrong, the prediction is semantically close to the actual ground truth which shows up as second prediction of the network.}

\end{figure}

The summary of training and testing cross entropy losses, percent correct accuracy and entropy of softmax predictions are tabulated in Table \ref{table3}. Deep CNN achieved high test accuracy of $98.765\%$. The entropy of softmax outputs of training images is $9.79495$. This indicates that the network is very confident not only when it is making correct predictions but also when it is making wrong predictions. An example of DeepCNN making very confident predictions can be seen in Figure \ref{fig:softmaxpreds} (a) and (b).

\begin{figure}[tpb]
\label{fig:wrong_tests_samples}
      \centering
	  {\includegraphics[width=\linewidth,keepaspectratio]{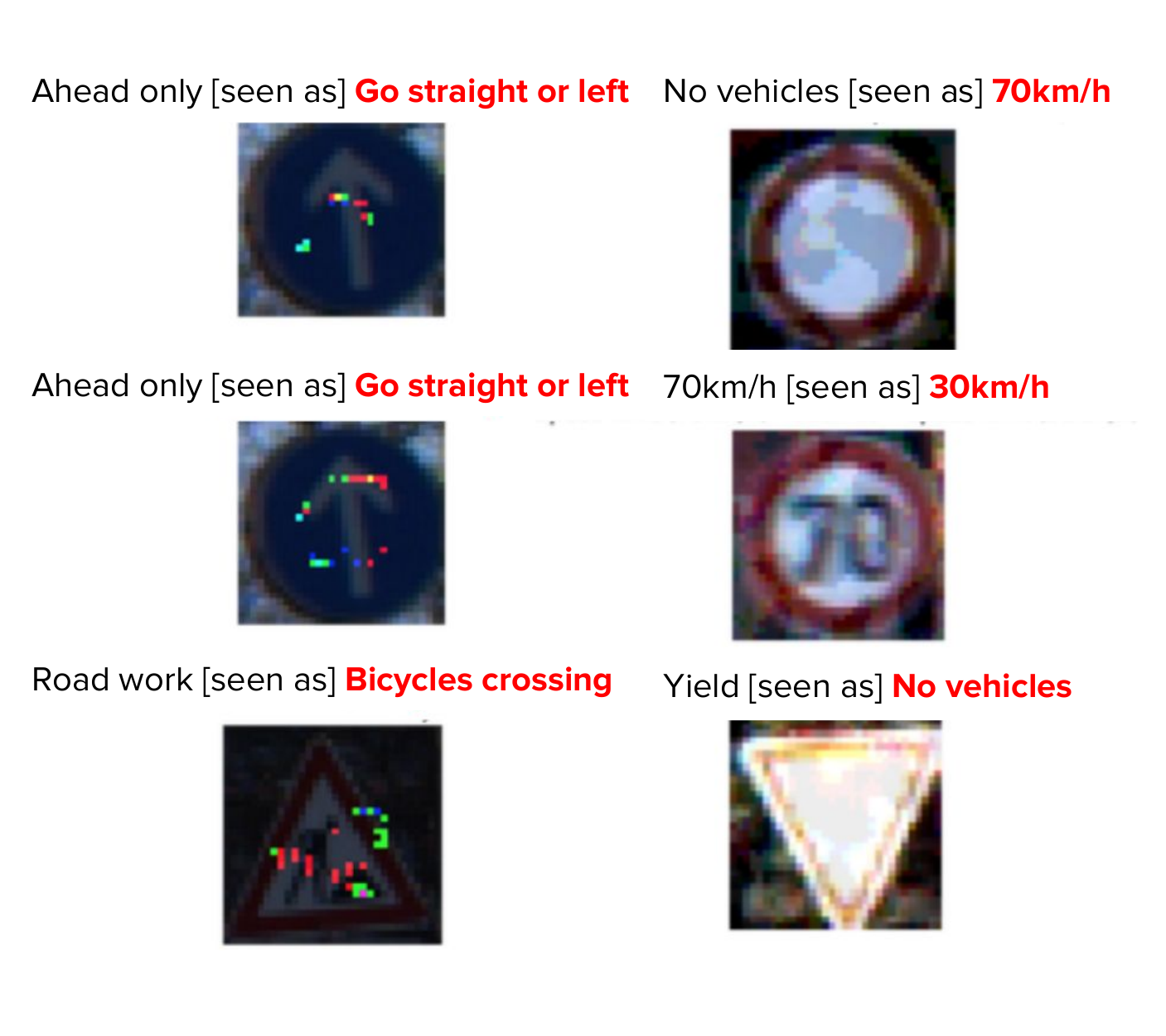}}
      \caption{Adversarial samples successfully fooled the DeepCNN to misclassify. Left column: adversarial samples with Jacobian Saliency Map Method. Right column: adversarial samples with Fast Gradient Sign Method.  The original samples of these road signs were correctly classified by DeepCNN.}
      \label{fig:wrong_test_samp} 
\end{figure}

Then we used adversarial examples generated 
with Fast Gradient Sign Method and Jacobian Saliency Map Method to test our DeepCNN. The test accuracy went down to $50.8859\%$ and cross entropy loss rose up to $3.68216$. Table \ref{table4} summarizes the difference between the test statistics on legitimates samples and adversarial samples. Figure \ref{fig:wrong_test_samp} shows some samples of road signs which were initially correctly classified but were misclassified when they are perturbed with Fast Gradient Sign Method and Jacobian Saliency Map Attack.

This indicates that adversarial examples generate by different network (AdversarialCNN) with no knowledge of the architecture and weights of another network (DeepCNN), can still attack the network up to a certain extent. The only shard knowledge between two networks is training data and training labels. This signifies our primary purpose of trying to attack Deep Neural Networks for a particular classification works without knowing the architecture, weights and parameters of that network.

\subsection{Defensive Distillation and Adversarial Training}

In order to train DeepCNN with defensive distillation, temperature parameter $T$ is set to $100$ which forces the network to produce more entropic outputs. An example of the softmax outputs of distilled network can be seen in Figure \ref{fig:softmaxpreds} (c) and (d). When trained with defensive distillation, even when the network is making a mistake (Figure \ref{fig:softmaxpreds} (c)), it is not very far off from the correct prediction. Opposed to that when the network is trained without distillation, the network is very confident on the first prediction but the second prediction is semantically far away from the correct class (Figure \ref{fig:softmaxpreds} (a)).

After training DeepCNN with defensive distillation method, the network's output accuracy on test set went down (Table \ref{table3} $T=100$ column) but not considerably. But now the network is making more entropic outputs as it can be seen in Table \ref{table3} Entropy of Training Softmax predictions at $T=100$ as well as Figure \ref{fig:softmaxpreds} (c) and (d). This indicates that the network is less confident when making right predictions as well as less confident when it is making wrong predictions. 

Right after the network is trained with defensive distillation, the network is fed with adversarial training set. Using the adversarial training set, the network is trained once again with defensive distillation.

\begin{figure}[tpb]
      \centering
	  {\includegraphics[width=\linewidth]{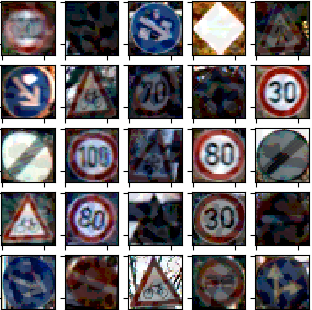}}
      \caption{Road sign image that were previously misclassified by DeepCNN but then correctly classified after defensive distillation and adversarial training.}
      \label{diff_mont} 
\end{figure}

When testing on adversarial samples, the deterioration of test accuracy and cross entropy loss is less dramatic then when compared to DeepCNN without defensive distillation and adversarial training. This effect can be seen in Table \ref{table5}. Some of the samples which were misclassifed by DeepCNN before defensive distillation and adversarial training but got correctly classified after defensive distillation and adversarial training can be seen in Figure \ref{diff_mont}.

Defensive distillation smoothen the model learned by a DeepCNN during training by helping the model generalize better to samples outside of its training dataset.
Distillation generates smoother classifier models by reducing their sensitivity to input perturbations. These smoother classifiers are found to be more resilient to adversarial samples and have improved class generalizability properties. \cite{papernot2016distillation}

\section{Discussion}
\label{sec:discussion}
The methods of crafting adversarial examples described in the paper is successful in generating a black box attack on a Deep neural network. The disadvantages of our method is the absence of a measurement parameter user perceptibility. While parameters such as $\epsilon$ from FGSM and distortion rate from Jacobian Saliency can be chosen, the perceptibility depends on the original image itself. Extensive studies on parameterizing the perceptibility of adversarial perturbation and increasing/decreasing the feature vector size of the images can be explored in future work. 

Furthermore, more comparisons can be made on visual differences of the FGSM method and the Jacobian saliency map method. The first method applies minuscule perturbations to almost all the pixels in the image while the latter applies very distinct perturbation to only a small amount of pixels. This can either be advantageous or disadvantageous depending on the situation. For example in \cite{evtimov2017robust}, the authors used more distinct modifications for physical adversarial examples as it is easier to apply. 

Defense against adversarial samples are not analogous to reducing the variance of the mode by regularization because adversarial examples are not traditional overfitting problem \cite{goodfellow2014explaining}.
Previous work has showed that a wide variety of traditional regularization methods including dropout and weight decay either fail to defend against adversarial examples or only do so by seriously harming accuracy on the original task.

Another weakness of making more robust neural networks in our work is that after defensive distillation and adversarial training, the test accuracy cannot reach the original test accuracy on non-adversarial samples. Moreover, early stopping with respect to a threshold parameter have affects on how well defensive distillation can work on adversarial samples. 

Since the data that we used for our work is cropped 32x32 images, much research should be done when the input is the image of the whole scene and only some perturbations are done around and on the object of interest in the scene.

\begin{table}[t]

\centering
\caption{DeepCNN trained without defensive distillation $(T=1)$ and with defensive distillation $(T=100)$. Temperature parameter T is tuned to give the best test accuracy while ensuring that the softmax predictions are as entropic as possible.}
\label{table3}
\begin{tabular}{|l|r|r|}
\hline
\multicolumn{1}{|c|}{}                                                            & \textbf{T = 1}     & \textbf{T = 100}   \\ \hline
Training Cross Entropy Loss                                                       & 0.00453   & 0.20243   \\ \hline
Training Accuracy                                                                 & 99.8766\% & 99.4563\% \\ \hline
Testing Cross Entropy Loss                                                        & 0.05645   & 0.20855   \\ \hline
Testing Accuracy                                                                  & 98.7650\% & 98.2999\% \\ \hline
Testing F1 Score                                                                  & 0.98765 & 0.98299 \\ \hline
\begin{tabular}[c]{@{}l@{}}Entropy of Training Softmax\\ predictions\end{tabular} & 9.79495   & 10.34862  \\ \hline
\end{tabular}
\bigskip
\centering
\caption{DeepCNN on legitimate test samples and on adversarial samples without defensive distillation. Notice that the test accuracy and test cross entropy loss deteriorate dramatically when tested on adversarial examples generated by AdversarialCNN.}
\label{table4}
\begin{tabular}{|l|r|r|}
\hline
\multirow{2}{*}{}                                                                 & \multicolumn{2}{c|}{\begin{tabular}[c]{@{}c@{}}Without Defensive Distillation and \\ Adversarial Training\end{tabular}} \\ \cline{2-3} 
                                                                                  & \multicolumn{1}{c|}{On legit samples}                   & \multicolumn{1}{c|}{On adversarial samples}                   \\ \hline
\begin{tabular}[c]{@{}l@{}}Testing Cross \\ Entropy loss\end{tabular}             & 0.05645                                                 & 3.68216                                                       \\ \hline
Testing Accuracy                                                                 & 98.7650\%                                               & 50.8859\%                                                     \\ \hline
Testing F1 Score                                                                 & 0.98765                                               & 0.50886                                                     \\ \hline

\begin{tabular}[c]{@{}l@{}}Entropy of Testing \\ Softmax predictions\end{tabular} & 8.77761                                                 & 8.42678                                                       \\ \hline
\end{tabular}
\bigskip
\centering
\caption{DeepCNN on legitimate test samples and on adversarial samples with defensive distillation with T=100. Notice that the test accuracy and test cross entropy loss deteriorate less dramatically compared to the network without defensive distillation}
\label{table5}
\begin{tabular}{|l|r|r|}
\hline
\multirow{2}{*}{}                                                                 & \multicolumn{2}{c|}{\begin{tabular}[c]{@{}c@{}}With Defensive Distillation and \\ Adversarial Training\end{tabular}} \\ \cline{2-3} 
                                                                                  & \multicolumn{1}{c|}{On legit samples}                  & \multicolumn{1}{c|}{On adversarial samples}                 \\ \hline
\begin{tabular}[c]{@{}l@{}}Testing Cross \\ Entropy loss\end{tabular}             & 0.20855                                                & 0.40542                                                     \\ \hline
Testing Accuracy                                                                 & 98.2999\%                                              & 91.4648\%                                                   \\ \hline
Testing F1 Score                                                                 & 0.98299                                              & 0.91463                                                   \\ \hline
\begin{tabular}[c]{@{}l@{}}Entropy of Testing \\ Softmax predictions\end{tabular} & 9.23198                                                & 10.40873                                                    \\ \hline
\end{tabular}
\end{table}

\section{Conclusion}
\label{sec:conclusion}
In this work, we have built a complete pipeline for crafting adversarial samples, building a deep Convolutional Neural Network which achieves a near state-of-the-art test set accuracy on GTSRB dataset, using the adversarial examples to attack a near state-of-the-art network without any prior knowledge of its architecture and building a more robust neural network to be more resilient to those adversarial examples. We have successfully crafted adversarial samples on GTSRB dataset using Fast Gradient Sign Method and Jacobian Saliency Map Method. Using those adversarial samples, we were able to fool a different network in a black box manner since we did not have access to the internal weights and architecture of that network. Finally we have successfully made that network more robust by defensive distillation and adversarial training. Our work serves as an aggregated implementation of work done by previous literature on adversarial sample crafting and defense against adversarial attack methods. Our work also differs form previous work on adversarial sample crafting and defense against adversarial samples in a way that we are aggregating, applying and verifying the previous methods in the context of traffic signs rather than on MNIST or on CIFAR10 which is mostly used by previous literature \cite{goodfellow2014explaining} \cite{papernot2016distillation} \cite{papernot2016limitations}.
\bibliography{references}

\begin{thebibliography}{10}

\bibitem{goodfellow2014explaining}
Ian~J Goodfellow, Jonathon Shlens, and Christian Szegedy.
\newblock Explaining and harnessing adversarial examples.
\newblock {\em arXiv preprint arXiv:1412.6572}, 2014.

\bibitem{papernot2016limitations}
Nicolas Papernot, Patrick McDaniel, Somesh Jha, Matt Fredrikson, Z~Berkay
  Celik, and Ananthram Swami.
\newblock The limitations of deep learning in adversarial settings.
\newblock In {\em Security and Privacy (EuroS\&P), 2016 IEEE European Symposium
  on}, pages 372--387. IEEE, 2016.

\bibitem{papernot2016distillation}
Nicolas Papernot, Patrick McDaniel, Xi~Wu, Somesh Jha, and Ananthram Swami.
\newblock Distillation as a defense to adversarial perturbations against deep
  neural networks.
\newblock In {\em Security and Privacy (SP), 2016 IEEE Symposium on}, pages
  582--597. IEEE, 2016.

\bibitem{nguyen2015deep}
Anh Nguyen, Jason Yosinski, and Jeff Clune.
\newblock Deep neural networks are easily fooled: High confidence predictions
  for unrecognizable images.
\newblock In {\em Proceedings of the IEEE Conference on Computer Vision and
  Pattern Recognition}, pages 427--436, 2015.

\bibitem{szegedy2013intriguing}
Christian Szegedy, Wojciech Zaremba, Ilya Sutskever, Joan Bruna, Dumitru Erhan,
  Ian Goodfellow, and Rob Fergus.
\newblock Intriguing properties of neural networks.
\newblock {\em arXiv preprint arXiv:1312.6199}, 2013.

\bibitem{papernot2016practical}
Nicolas Papernot, Patrick McDaniel, Ian Goodfellow, Somesh Jha, Z~Berkay Celik,
  and Ananthram Swami.
\newblock Practical black-box attacks against deep learning systems using
  adversarial examples.
\newblock {\em arXiv preprint arXiv:1602.02697}, 2016.

\bibitem{kurakin2016adversarial}
Alexey Kurakin, Ian Goodfellow, and Samy Bengio.
\newblock Adversarial examples in the physical world.
\newblock {\em arXiv preprint arXiv:1607.02533}, 2016.

\bibitem{evtimov2017robust}
Ivan Evtimov, Kevin Eykholt, Earlence Fernandes, Tadayoshi Kohno, Bo~Li, Atul
  Prakash, Amir Rahmati, and Dawn Song.
\newblock Robust physical-world attacks on machine learning models.
\newblock {\em arXiv preprint arXiv:1707.08945}, 2017.

\bibitem{lu2017no}
Jiajun Lu, Hussein Sibai, Evan Fabry, and David Forsyth.
\newblock No need to worry about adversarial examples in object detection in
  autonomous vehicles.
\newblock {\em arXiv preprint arXiv:1707.03501}, 2017.

\bibitem{lu2017standard}
Jiajun Lu, Hussein Sibai, Evan Fabry, and David Forsyth.
\newblock Standard detectors aren't (currently) fooled by physical adversarial
  stop signs.
\newblock {\em arXiv preprint arXiv:1710.03337}, 2017.

\bibitem{ren2015faster}
Shaoqing Ren, Kaiming He, Ross Girshick, and Jian Sun.
\newblock Faster r-cnn: Towards real-time object detection with region proposal
  networks.
\newblock In {\em Advances in neural information processing systems}, pages
  91--99, 2015.

\bibitem{redmon2016yolo9000}
Joseph Redmon and Ali Farhadi.
\newblock Yolo9000: better, faster, stronger.
\newblock {\em arXiv preprint arXiv:1612.08242}, 2016.

\bibitem{hinton2015distilling}
Geoffrey Hinton, Oriol Vinyals, and Jeff Dean.
\newblock Distilling the knowledge in a neural network.
\newblock {\em arXiv preprint arXiv:1503.02531}, 2015.

\bibitem{carlini2016defensive}
Nicholas Carlini and David Wagner.
\newblock Defensive distillation is not robust to adversarial examples.
\newblock {\em arXiv preprint arXiv:1607.04311}, 2016.

\bibitem{Stallkamp-IJCNN-2011}
Johannes Stallkamp, Marc Schlipsing, Jan Salmen, and Christian Igel.
\newblock The {G}erman {T}raffic {S}ign {R}ecognition {B}enchmark: A
  multi-class classification competition.
\newblock In {\em IEEE International Joint Conference on Neural Networks},
  pages 1453--1460, 2011.

\bibitem{wangeffectiveness}
Jason Wang and Luis Perez.
\newblock The effectiveness of data augmentation in image classification using
  deep learning.

\bibitem{paper2016jsma}
Somesh Jha Matt Fredrikson Z. Berkay Celik Ananthram~Swami Nicolas~Papernot,
  Patrick~McDaniel.
\newblock The limitations of deep learning in adversarial settings.
\newblock In {\em Security and Privacy (SP), 2016 IEEE Symposium on}. IEEE,
  2016.

\bibitem{papernot2017cleverhans}
Ian Goodfellow Reuben Feinman Fartash Faghri Alexander Matyasko Karen
  Hambardzumyan Yi-Lin Juang Alexey Kurakin Ryan Sheatsley Abhibhav Garg
  Yen-Chen~Lin Nicolas~Papernot, Nicholas~Carlini.
\newblock cleverhans v2.0.0: an adversarial machine learning library.
\newblock {\em arXiv preprint arXiv:1610.00768}, 2017.

\end{thebibliography}
\bibliographystyle{unsrt}

\end{document}